\documentclass[runningheads]{llncs}
\usepackage[T1]{fontenc}
\usepackage{graphicx}
\usepackage{cite}
\usepackage{amsmath}
\usepackage{amssymb}
\usepackage{booktabs}
\usepackage{algorithm}
\usepackage{algorithmic}
\usepackage{url}

\begin{document}

\title{Unsupervised State Representation Learning in Partially~Observable~Atari~Games}

\author{Li Meng\inst{1}\orcidID{0000-0002-8867-9104} \and
Morten Goodwin\inst{2,3}\orcidID{0000-0001-6331-702X} \and
Anis Yazidi\inst{3}\orcidID{0000-0001-7591-1659} \and
 Paal Engelstad\inst{1}\orcidID{0009-0000-8371-927X}}

\authorrunning{L. Meng et al.}
\titlerunning{Unsupervised State Representation Learning in Games}
\institute{ University of Oslo, Norway, \and University of Agder, Norway \and
 Oslo Metropolitan University, Norway}
\maketitle

\begin{abstract}
State representation learning aims to capture latent factors of an environment. Contrastive methods have performed better than generative models in previous state representation learning research. Although some researchers realize the connections between masked image modeling and contrastive representation learning, the effort is focused on using masks as an augmentation technique to represent the latent generative factors better. Partially observable environments in reinforcement learning have not yet been carefully studied using unsupervised state representation learning methods.

In this article, we create an unsupervised state representation learning scheme for partially observable states. We conducted our experiment on a previous Atari 2600 framework designed to evaluate representation learning models. A contrastive method called Spatiotemporal DeepInfomax (ST-DIM) has shown state-of-the-art performance on this benchmark but remains inferior to its supervised counterpart. Our approach improves ST-DIM when the environment is not fully observable and achieves higher F1 scores and accuracy scores than the supervised learning counterpart. The mean accuracy score averaged over categories of our approach is  \(\sim \) 66 \%, compared to \(\sim \) 38 \% of supervised learning. The mean F1 score is  \(\sim \) 64 \% to \(\sim \) 33 \%. The code can be found on \url{https://github.com/mengli11235/MST_DIM}.


\end{abstract}

\section{Introduction}
Deep representation learning is a machine learning (ML) type that focuses on learning useful data representations. These representations can be learned using deep neural networks (NNs) and transferred to a variety of downstream computer vision (CV), and natural language processing (NLP) tasks \cite{doersch2017multi, kolesnikov2019revisiting}. Deep representation learning includes autoencoders \cite{kingma2013auto}, generative models \cite{gregor2015draw}, contrastive methods \cite{oord2018representation, he2020momentum} and transformer models \cite{devlin2018bert}. 

State representation learning (SRL) \cite{jonschkowski2015learning, lesort2018state, anand2019unsupervised} is a particular field of representation learning where the state observations are commonly seen in the reinforcement learning (RL) setup. Agents can interact with the environment, which itself changes accordingly throughout interactions. RL is a well-established ML field that solves the Markov decision process (MDP) \cite{puterman1990markov}. Traditional RL methods such as Q-learning \cite{watkins1992q} have evolved through adopting NNs \cite{mnih2013playing}. Moreover, convolutional neural networks (CNNs) are deployed in environments with image inputs, and RL agents can learn from raw pixels.

Partially observable Markov decision processes (PODMPs) \cite{cassandra1994acting, sutton2018reinforcement} are MDPs where an agent can only observe a limited part of the environment, not the full state. Recently, there have been some developments in decoupling SRL from RL \cite{lee2020predictive, laskin2020curl, guo2020bootstrap, stooke2021decoupling}. More improvements by using SRL have been reported in partially observable environments than in fully observable ones. However, it is not clear how the POMDP state is captured and preserved by representations.

This paper designs an unsupervised representation learning scheme for partially observable environments. This method extends ST-DIM \cite{anand2019unsupervised} and introduces an unsupervised pretraining setting suitable to partially observable Atari Games. Different pretraining hyper-parameter choices are also discussed in our ablation study.

Our contribution is summarized as follows: (1) We propose MST-DIM, a contrastive method suitable to pretrain data collected in a partially observable environment. (2) We test our method on the SRL benchmark using 20 Atari 2600 games and compare the results with the ST-DIM and supervised methods. (3) Should the percentages of the observable parts of states be the same in pretraining and in probing? Extensive evaluations are conducted to examine what is needed for SRL in order to let the model accurately predict the ground truth labels.

\section{Related Work}
\noindent \textbf{Self-supervised Learning} Self-supervised learning learns useful representations from unlabeled data, which can be used in various downstream tasks. The methodology has played an important role in NLP \cite{devlin2018bert} and CV fields. Contrastive Predictive Coding (CPC) \cite{oord2018representation} learns predictive representations by capturing the information that is maximally useful to predict future (spatial or temporal) samples. SimCLR \cite{chen2020simple} provides a simple yet effective framework for contrastive learning. Momentum Contrast (MoCo) keeps dynamic dictionaries for contrastive learning. Self-supervised Vision Transformers \cite{chen2021empirical} study the usage of ViTs (\cite{dosovitskiy2020image}) on Momentum Contrast (MoCo) \cite{he2020momentum}.

\noindent \textbf{Contrastive Representations for RL} Contrastive Unsupervised Representations for RL (CURL) \cite{laskin2020curl} is an RL pipeline that extracts high-level features using an auxiliary contrastive loss, which can be combined with on-policy or off-policy RL algorithms. Masked Contrastive Representation Learning (M-CURL) \cite{zhu2022masked} improves the data efficiency in CURL by considering the correlation among consecutive inputs and using masks to help the transformer module learn to reconstruct the features of the ground truth. The loss is defined as a sum of the RL loss and masked contrastive loss, and the transformer module is discarded during inference. M-CURL reportedly outperforms CURL on 21 out of 26 environments from Atari 2600 Games.  

\noindent \textbf{Representation Learning in POMDPs} Predictions of Bootstrapped Latents (PBL) \cite{guo2020bootstrap} is an RL algorithm designed for the multitask setting. PBL is trained by predicting future latent observations from partial histories and the current states from latent observations. Histories in POMDPs are typically compressed into a current agent state using NNs \cite{sutton2018reinforcement}. Agents trained with PBL achieve significantly higher human normalized scores than baseline methods in the partially observable DMLab 30 environments, but the gap is at most minimal in fully observable environments. Augmented Temporal Contrast (ATC) \cite{stooke2021decoupling} associates temporally close pairs of observations and also shows its usefulness in POMDPs.

\section{Method}
ST-DIM \cite{anand2019unsupervised} is a method that captures the latent generative factors through maximizing mutual information lower-bound estimate over consecutive
observations $x_t$ and $x_{t+1}$ given a set of cross-episode observations $\chi = \{x_1,x_2,...,x_n\}$, originated from agents interacting with RL environments. It uses infoNCE \cite{oord2018representation} that maximizes Eq. \ref{eq:infonce} as the mutual information estimator between patches as Deep InfoMax (DIM)\cite{hjelm2018learning} does.

\begin{equation}
\scriptsize
\mathcal{I}_{NCE}(\{(x_i, y_i)\}^N_{i=1})=\sum^{N}_{i=1}log\frac{exp\;f(x_i,y_i)}{\sum^{N}_{j=1}exp\;f(x_i,y_j)}
\label{eq:infonce}
\end{equation}

For any i, $(x_i, y_i)$ is called positive examples from the joint distribution $p(x,y)$ and  $(x_i, y_j)$ from the product of marginals $p(x)p(y)$ is called negative examples for any $i\neq j$. Meanwhile, $f(x,y)$ is a score function, i.e., a bilinear layer.

\begin{figure}[t]
    \centering
    \includegraphics[width=0.4\linewidth]{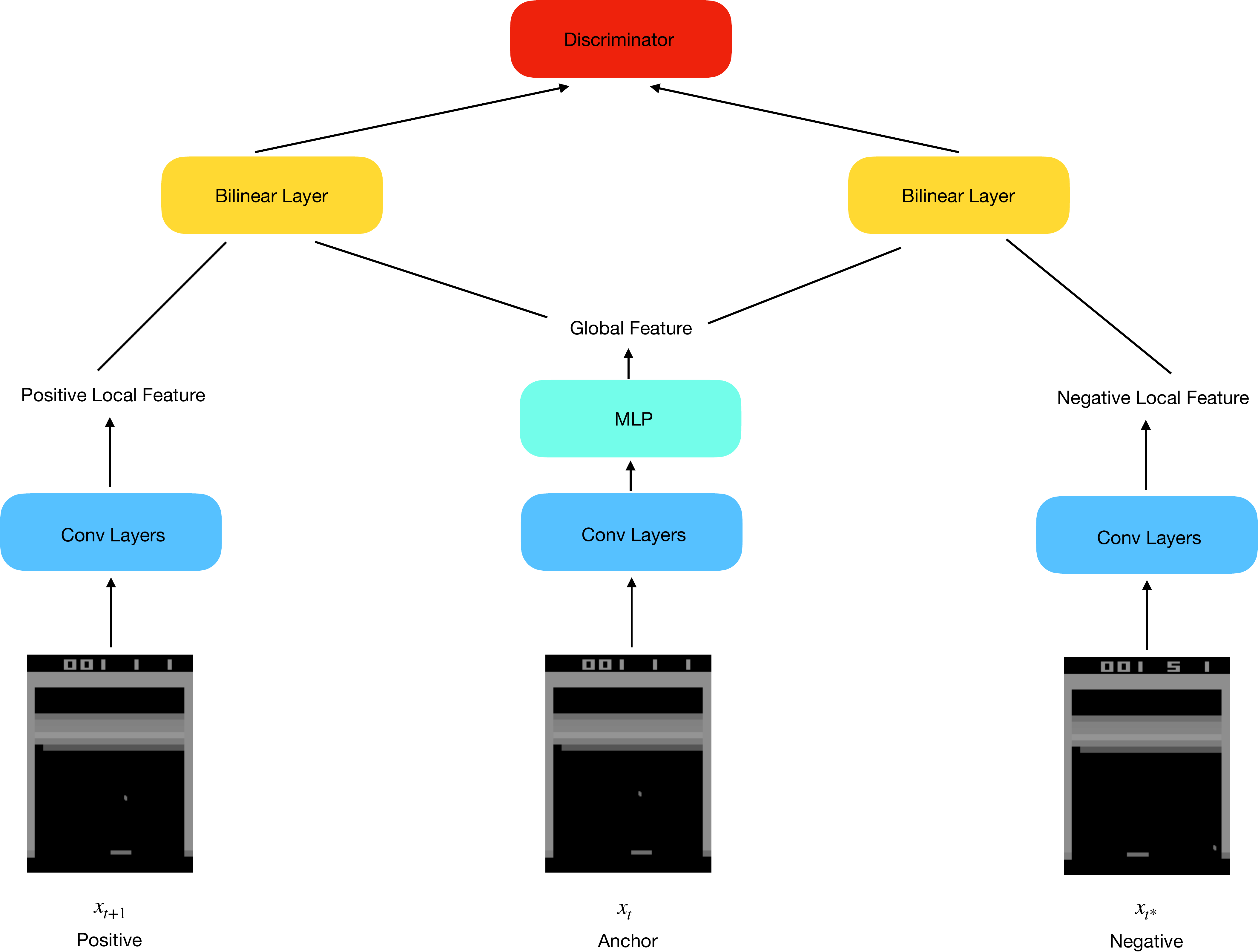}
    \caption{An illustration of the global-local contrastive task in ST-DIM. For the local-local contrastive task, we discard the MLP and use the local feature of the anchor.}
\label{fig:stdim}
\end{figure}

ST-DIM utilizes both the global-local (Eq. \ref{eq:gl}) and local-local objective (Eq. \ref{eq:ll}). An illustration of the global-local contrastive task is also shown in Figure \ref{fig:stdim}. The difference between the local-local and global-local tasks is that an additional MLP is used to extract the global features.

\begin{equation}
\scriptsize
\mathcal{L}_{GL}=\sum^{M}_{m=1}\sum^{N}_{n=1}-log\frac{exp\;g_{m,n}(x_t,x_{t+1})}{\sum_{x_{t*} \in X_{next}}exp\;g_{m,n}(x_t,x_{t*})}
\label{eq:gl}
\end{equation}

\begin{equation}
\scriptsize
\mathcal{L}_{LL}=\sum^{M}_{m=1}\sum^{N}_{n=1}-log\frac{exp\;f_{m,n}(x_t,x_{t+1})}{\sum_{x_{t*} \in X_{next}}exp\;f_{m,n}(x_t,x_{t*})}
\label{eq:ll}
\end{equation}

Here, $M$ and $N$ are the height and width, $g_{m,n}(x_t, x_{t+1}) = \phi(x_t)^T W_g\phi_{m,n}(x_{t+1})$ and $\phi_(m,n)$ is the local feature vector produced by convolutional layers in the representation encoder $\phi$ at the location $(m, n)$. On the other hand, $f_{m,n}(x_t, x_{t+1}) = \phi_{m,n}(x_t)^T W_l\phi_{m,n}(x_{t+1})$. Observations $x_t$ and $x_{t+1}$ are temporally adjacent, whereas $x_{t*}$ is a randomly sampled observation from the minibatch.

In order to fit ST-DIM into partially observable environments, we propose MST-DIM and define random masks $k_t$ for each consecutive pair $(x_t,x_{t+1})$ that is drawn from a binomial distribution $\mathbb{K}$. Therefore, Eq. \ref{eq:gl} and Eq. \ref{eq:ll} are modified as Eq. \ref{eq:mgl} and Eq. \ref{eq:mll}:

\begin{equation}
\scriptsize
\mathcal{L}_{MGL}=\sum^{M}_{m=1}\sum^{N}_{n=1}-log\frac{exp\;g_{m,n}(x_tk_t,x_{t+1})}{\sum_{x_{t*} \in X_{next}}exp\;g_{m,n}(x_tk_t,x_{t*})}
\label{eq:mgl}
\end{equation}

\begin{equation}
\scriptsize
\mathcal{L}_{MLL}=\sum^{M}_{m=1}\sum^{N}_{n=1}-log\frac{exp\;f_{m,n}(x_tk_t,x_{t+1})}{\sum_{x_{t*} \in X_{next}}exp\;f_{m,n}(x_tk_t,x_{t*})}
\label{eq:mll}
\end{equation}

Meanwhile, we can tune the probability of masking in $\mathbb{K}$ for both the pretraining and probing. For example, a masking ratio of $0.4$ means that 40 \% of the full observation is not visible to the agent.

\section{Experimental Details}

Our experiment is conducted among 20 Atari games of Arcade Learning Environment (ALE). Performances are evaluated by probe accuracy and F1 scores for each game. Because ALE does not directly provide ground truth information, ST-DIM has been conducted on the newly designed Atari Annotated RAM Interface (AtariARI) \cite{anand2019unsupervised} that exposes the state variables from the source code \cite{taylor2008playing} in 22 games. State variables are categorized as agent localization (Agent Loc.), small object localization (Small Loc.), other localization (Other Loc.), score/clock/lives/display, and miscellaneous (Misc.). Detailed descriptions of states for each game across categories can be found in the original paper \cite{anand2019unsupervised}. We also summarize probe accuracy and F1 scores across those state categories for each game in our experiment. Not all categories are available for each game, and we only include results from applicable ones.

Due to practical implementation issues, we exclude Berzerk, Riverraid and Yars Revenge and include Battle Zone in our experiment, making a total number of 20 games. Trajectories collected by random agents are used in our experiment, as ST-DIM \cite{anand2019unsupervised} suggested it can be a better choice than using trajectories from PPO agents.

\begin{figure}[t]
    \centering
    \includegraphics[width=0.8\linewidth]{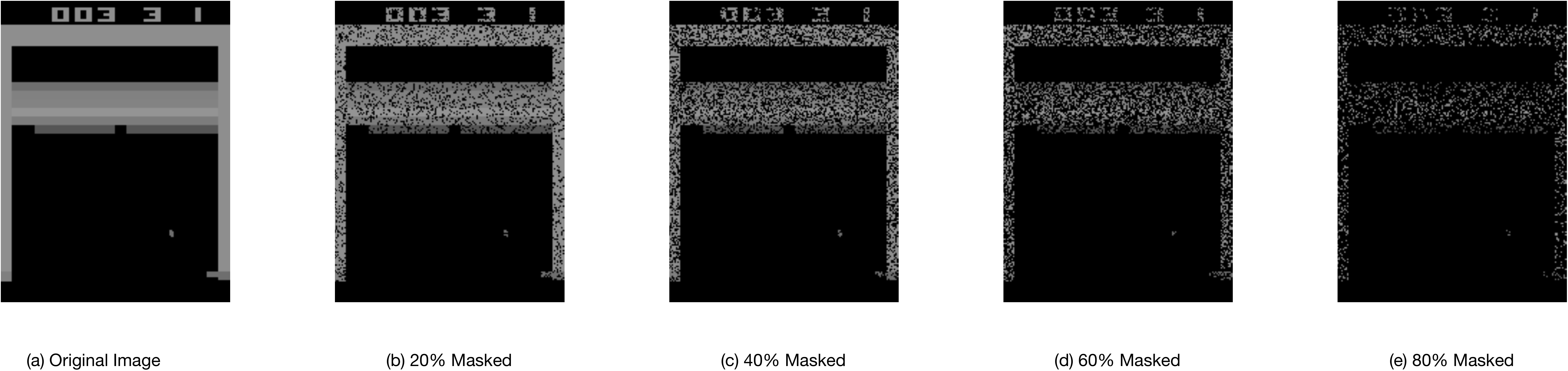}
    \caption{The original image is masked by different percentages of random noise.}
\label{fig:masked}
\end{figure}

For pretraining, we use different partially observable setups. To verify if the masking ratio in pretraining can be different from the probing, five types of pretraining images have been considered in our experiment, as illustrated by Figure \ref{fig:masked}. The images can be original, 20\% masked, 40\% masked, 60\% masked, or 80\% masked.

\begin{figure}[t]
    \centering
    \includegraphics[width=0.4\linewidth]{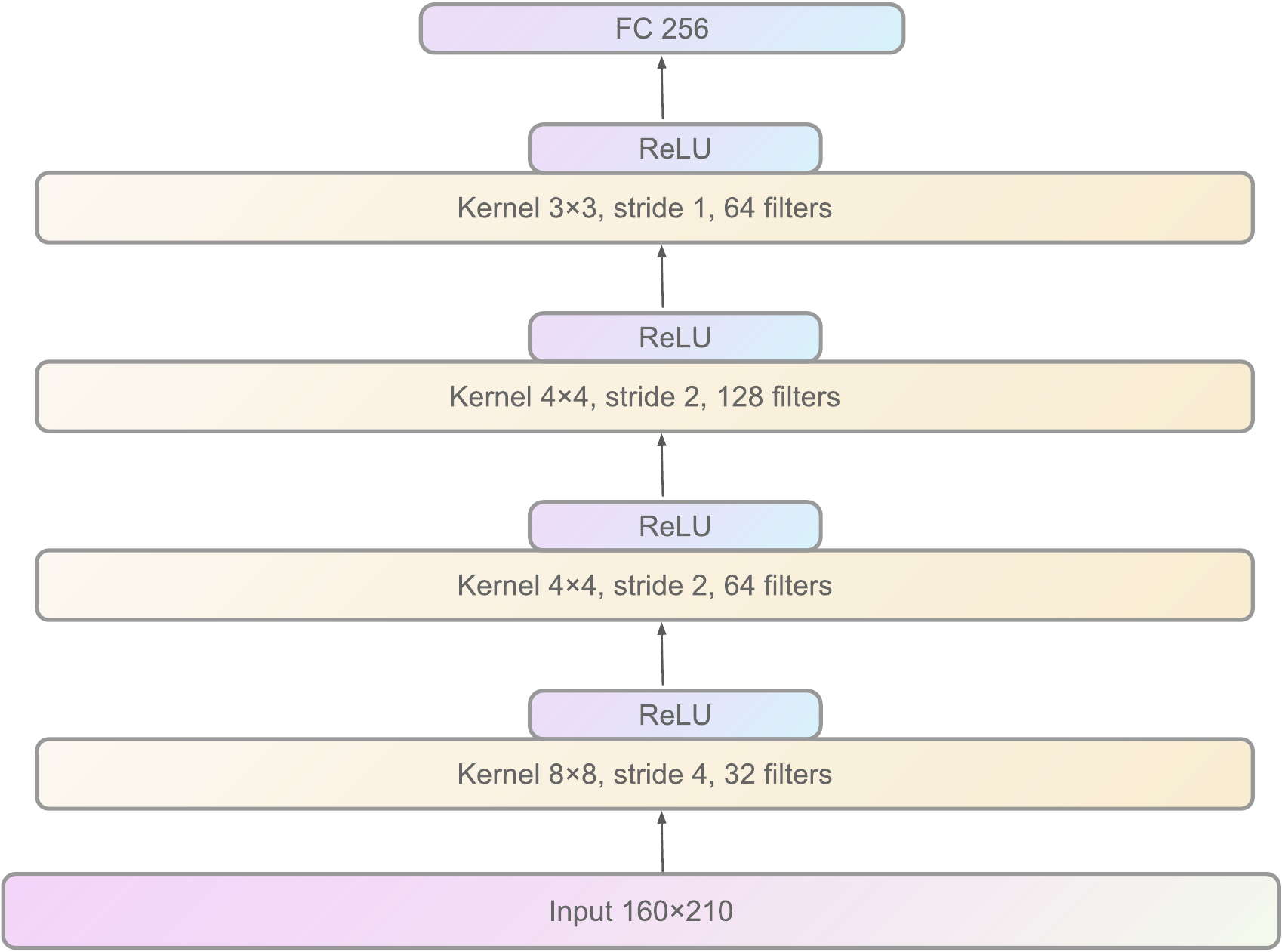}
    \caption{Architecture of the encoder.}
\label{fig:nn}
\end{figure}

We follow the same probing protocol as ST-DIM and focus on the explicitness, i.e., how well the latent generative factors can be recovered. This is done by training a linear classifier that predicts the state variables using the learned representations. We keep the hyper-parameters the same as ST-DIM to make our experiment comparable. A short list of hyper-parameters is shown in Table \ref{tbl:para}. Setting the entropy threshold removes large objects that have low entropy from the labels. The encoder architecture is the same as in ST-DIM, as illustrated by Figure \ref{fig:nn}.

\begin{table}[t]
\scriptsize
\caption{Parameter Choices}
\centering
\begin{tabular}{c c} 
\hline
\textbf{Hyper-parameter} & \textbf{Value} \\
Image Size & $160\times210$ \\
Batch Size & 64\\
Learning Rate & 3e-4\\
Entropy Threshold & 0.6\\
Pretraining Steps & 80000\\
Probe Training Steps & 35000\\
Probe Testing Steps & 10000\\
\hline
\end{tabular}
\label{tbl:para}
\end{table}

\section{Results}
In this section, we demonstrate the performance of unsupervised representation learning in the partially observable reinforcement learning environment. Table \ref{tbl:res1} shows the F1 scores for each game and Table \ref{tbl:res2} shows the accuracy. "Observable" represents the setting where the probing is implemented with full observations.  "Non-Observable" is the ST-DIM setting where the pretraining is with full observations, but the probing are with partial observations. "Supervised" is the setting where no pretraining is included but the model is trained and tested only in probing using the supervised manner. "Ratio" indicates the masking ratio in probing, e.g., ratio 0.2 equals that 20\% part of each observation in probing is not visible. By default, the masking ratio in probing is set to 0.4.

It is obvious that the model is capable of predicting the state variables the most in fully observable environments, as the observations in probing are not masked by noise. The rest results are all from models trained and tested with masked observations for probing. The performance deteriorates considerably if the model still uses ST-DIM and pretrains on full observations. Supervised training also exhibits the same characteristic and obtains similar results with ST-DIM. On the other hand, MST-DIM achieves significantly higher accuracy and F1 scores by taking advantage of masked pretraining. Pretraining with a different masking ratio has also shown improvements over ST-DIM. Using masking ratios 0.2 and 0.6 in pretraining yields slightly inferior results than the default masking ratio of 0.4. Surprisingly, using a masking ratio of 0.8 still improves the ST-DIM despite most of the images being masked in this setting.

The mean accuracy score (0.66) and F1 score (0.64) of MST-DIM are slightly worse than the accuracy score (0.71) and F1 score (0.7) under the fully observable setup. However, these results considerably exceed those of supervised learning (0.38 and 0.33) and ST-DIM (0.38 and 0.34) under the same partially observable setup. For the other three masking ratios, a ratio of 0.6 achieves the highest accuracy (0.63) and F1 scores (0.61), a ratio of 0.2 achieves slightly lower scores (0.61 and 0.59), and a ratio of 0.8 obtains the worst accuracy and F1 scores (0.53 and 0.5).

Meanwhile, Table \ref{tbl:res3} and Table \ref{tbl:res4} show the results for each category averaged over games. It is clear that MST-DIM still performs the best category-wise in partially observable environments, and achieves scores that is slightly worse than ST-DIM that is probed under fully observable environments. A different masking ratio in pretraining still enhances the model's capability across games in probing tasks, and even a masking ratio (0.8) that is remote from the probing masking ratio (0.4) can facilitate achieving better scores for each ground truth category.


\begin{table}[t]
\scriptsize
\caption{Probe F1 scores of each game averaged across categories}
\centering
\begin{tabular}{c c | c c|c c c c} 
\hline
Games&Observable&Non-observable&Supervised&Pretrain&Ratio 0.2&Ratio 0.6&Ratio 0.8\\
\hline
Asteroids&\textbf{0.46}&0.39&0.39&\textbf{0.45}&\textbf{0.45}&0.44&0.44\\
\hline
Battle Zone&\textbf{0.5}&0.29&0.28&\textbf{0.45}&0.39&0.41&0.38\\
\hline
Bowling&\textbf{0.96}&0.29&0.29&\textbf{0.9}&0.72&0.85&0.63\\
\hline
Boxing&\textbf{0.59}&0.11&0.1&\textbf{0.53}&0.38&0.43&0.2\\
\hline
Breakout&\textbf{0.87}&0.37&0.37&\textbf{0.85}&0.83&\textbf{0.85}&0.29\\
\hline
Demon Attack&\textbf{0.66}&0.46&0.45&\textbf{0.64}&0.6&0.63&0.57\\
\hline
Freeway&\textbf{0.81}&0.03&0.03&\textbf{0.27}&\textbf{0.27}&0.1&0.05\\
\hline
Frostbite&\textbf{0.72}&0.33&0.34&\textbf{0.7}&0.65&0.66&0.58\\
\hline
Hero&\textbf{0.92}&0.57&0.58&\textbf{0.9}&0.87&0.88&0.84\\
\hline
Ms Pacman&\textbf{0.7}&0.35&0.36&\textbf{0.69}&0.64&0.68&0.62\\
\hline
Montezuma Revenge&\textbf{0.77}&0.54&0.53&\textbf{0.75}&0.74&0.74&0.71\\
\hline
Pitfall&\textbf{0.68}&0.24&0.25&\textbf{0.62}&0.53&0.6&0.54\\
\hline
Pong&\textbf{0.81}&0.13&0.13&\textbf{0.71}&0.69&0.65&0.42\\
\hline
Private Eye&\textbf{0.88}&0.5&0.49&\textbf{0.84}&0.81&0.82&0.68\\
\hline
Qbert&\textbf{0.72}&0.47&0.47&\textbf{0.71}&0.69&\textbf{0.71}&0.69\\
\hline
Seaquest&\textbf{0.64}&0.38&0.37&\textbf{0.62}&0.59&0.61&0.54\\
\hline
Space Invaders&\textbf{0.56}&0.45&0.44&0.56&0.55&\textbf{0.57}&0.56\\
\hline
Tennis&\textbf{0.6}&0.13&0.13&\textbf{0.48}&0.34&0.4&0.23\\
\hline
Venture&\textbf{0.55}&0.39&0.4&\textbf{0.54}&0.53&0.53&0.52\\
\hline
Video Pinball&\textbf{0.62}&0.29&0.3&0.62&0.57&\textbf{0.63}&0.6\\
\hline
Mean&\textbf{0.7}&0.34&0.33&\textbf{0.64}&0.59&0.61&0.5\\
\hline
\end{tabular}
\label{tbl:res1}
\end{table}

\begin{table}[t]
\scriptsize
\caption{Probe accuracy scores of each game averaged across categories}
\centering
\begin{tabular}{c c | c c|c c c c} 
\hline
Games&Observable&Non-observable&Supervised&Pretrain&Ratio 0.2&Ratio 0.6&Ratio 0.8\\\hline
Asteroids&\textbf{0.5}&0.46&0.46&\textbf{0.5}&\textbf{0.5}&\textbf{0.5}&0.49\\
\hline
Battle Zone&\textbf{0.52}&0.36&0.35&\textbf{0.48}&0.44&0.45&0.44\\
\hline
Bowling&\textbf{0.96}&0.36&0.36&\textbf{0.91}&0.73&0.85&0.66\\
\hline
Boxing&\textbf{0.59}&0.14&0.13&\textbf{0.54}&0.39&0.44&0.22\\
\hline
Breakout&\textbf{0.88}&0.41&0.41&\textbf{0.86}&0.84&\textbf{0.86}&0.37\\
\hline
Demon Attack&\textbf{0.66}&0.47&0.47&\textbf{0.65}&0.6&0.64&0.58\\
\hline
Freeway&\textbf{0.81}&0.06&0.06&\textbf{0.3}&\textbf{0.3}&0.15&0.09\\
\hline
Frostbite&\textbf{0.73}&0.38&0.38&\textbf{0.7}&0.66&0.67&0.59\\
\hline
Hero&\textbf{0.92}&0.59&0.59&\textbf{0.9}&0.88&0.88&0.84\\
\hline
Ms Pacman&\textbf{0.71}&0.4&0.4&\textbf{0.7}&0.66&0.69&0.65\\
\hline
Montezuma Revenge&\textbf{0.77}&0.55&0.54&\textbf{0.76}&0.74&0.74&0.72\\
\hline
Pitfall&\textbf{0.69}&0.3&0.3&\textbf{0.64}&0.55&0.62&0.57\\
\hline
Pong&\textbf{0.82}&0.21&0.21&\textbf{0.73}&0.7&0.67&0.47\\
\hline
Private Eye&\textbf{0.88}&0.52&0.51&\textbf{0.84}&0.81&0.83&0.68\\
\hline
Qbert&\textbf{0.73}&0.51&0.51&\textbf{0.72}&0.7&0.71&0.69\\
\hline
Seaquest&\textbf{0.66}&0.46&0.45&\textbf{0.63}&0.61&\textbf{0.63}&0.57\\
\hline
Space Invaders&\textbf{0.57}&0.48&0.47&\textbf{0.59}&0.57&\textbf{0.59}&0.58\\
\hline
Tennis&\textbf{0.61}&0.22&0.22&\textbf{0.51}&0.39&0.44&0.29\\
\hline
Venture&\textbf{0.56}&0.41&0.42&\textbf{0.55}&0.54&0.54&0.53\\
\hline
Video Pinball&\textbf{0.63}&0.31&0.32&0.62&0.57&\textbf{0.63}&0.61\\
\hline
Mean&\textbf{0.71}&0.38&0.38&\textbf{0.66}&0.61&0.63&0.53\\
\hline
\end{tabular}
\label{tbl:res2}
\end{table}

\begin{table}[t]
\scriptsize
\caption{Probe F1 scores of different ground truth categories averaged across all games}
\centering
\begin{tabular}{c c | c c|c c c c} 
\hline
Categories&Observable&Non-observable&Supervised&Pretrain&Ratio 0.2&Ratio 0.6&Ratio 0.8\\
\hline
Agent Loc. &\textbf{0.58}&0.26&0.26&\textbf{0.52}&0.48&0.5&0.41\\
\hline
Misc.&\textbf{0.73}&0.48&0.48&\textbf{0.72}&0.68&0.7&0.61\\
\hline
Other Loc.&\textbf{0.64}&0.34&0.34&\textbf{0.59}&0.56&0.54&0.47\\
\hline
Score/Clock/Lives/Display&\textbf{0.9}&0.42&0.42&\textbf{0.86}&0.77&0.83&0.7\\
\hline
Small Loc.&\textbf{0.53}&0.21&0.21&\textbf{0.47}&0.42&0.44&0.28\\
\hline
\end{tabular}
\label{tbl:res3}
\end{table}

\begin{table}[t]
\scriptsize
\caption{Probe accuracy scores of different ground truth categories}
\centering
\begin{tabular}{c c | c c|c c c c} 
\hline
Categories&Observable&Non-observable&Supervised&Pretrain&Ratio 0.2&Ratio 0.6&Ratio 0.8\\
\hline
Agent Loc. &\textbf{0.59}&0.32&0.32&\textbf{0.54}&0.5&0.52&0.44\\
\hline
Misc.&\textbf{0.74}&0.52&0.52&\textbf{0.73}&0.69&0.71&0.64\\
\hline
Other Loc.&\textbf{0.65}&0.38&0.38&\textbf{0.59}&0.58&0.55&0.49\\
\hline
Score/Clock/Lives/Display&\textbf{0.9}&0.44&0.44&\textbf{0.87}&0.78&0.84&0.71\\
\hline
Small Loc.&\textbf{0.55}&0.29&0.29&\textbf{0.5}&0.46&0.47&0.34\\
\hline
\end{tabular}
\label{tbl:res4}
\end{table}

\section{Discussion}

It was found that there was a sizable gap between the performances of ST-DIM and supervised training by \cite{anand2019unsupervised}. However, their difference is trivial under our partially observable setting. Meanwhile, MST-DIM has demonstrated better performance than both of them. The reason might be that ST-DIM and supervised methods do not possess better initializations, and yield similarly deteriorated results in partially observable environments.

On the other hand, Randomly initialized CNNs can perform reasonably well in probing tasks. Their scores are only slightly lower than those of generative methods in \cite{anand2019unsupervised}, because random CNNs are considered a strong prior in Atari games and can capture the inductive bias \cite{burda2018large, ulyanov2018deep}.

Different masking ratios in pretraining have shown to be effective, even for a large masking ratio that generates visually invisible images. Although the closer the pretraining masking ratio is, the more accurate the probing prediction can be, the masking ratio in pretraining is not required to be the same. The results indicate that unsupervised pretraining can learn reliable latent generative factors from a different data distribution. Unlike in fully observable environments, this is strong evidence that contrastive methods can play a key role in strengthening model capabilities in partially observable environments for downstream tasks.

The results of small object localization in Table \ref{tbl:res3} and \ref{tbl:res4} are highlighted in ST-DIM because generative methods typically do not penalize enough for not modeling the pixels making up small objects. The local-local contrastive task in ST-DIM is specialized in capturing local representation. In our experiment, it is clear that the performances of ST-DIM and the supervised method have dropped significantly in partially observable experiments because small objects 
 with few pixels can be easily masked out completely. On the other hand, MST-DIM overcomes this problem by masked pretraining and is close to achieving the same level of performance (0.47 to 0.53 for F1 scores and 0.5 to 0.55 in accuracy scores). However, MST-DIM has suffered from pretraining with a masking ratio of 0.8, which masks most of the image.

In some games, such as boxing, easy-to-learn features might saturate the objective and let contrastive methods fail. For example, contrastive methods other than ST-DIM fail to model features besides the clock in boxing \cite{anand2019unsupervised}. This is also a problem in partially observable environments and causes ST-DIM, supervised method, and MST-DIM with a masking ratio of 0.8 to perform worse. On the other hand, MST-DIM with a masking ratio close to 0.4 exhibits robustness as ST-DIM in fully observable environments.

For the study of different masking ratios, we observe that a masking ratio that is slightly higher than one of the underlying observations obtains the best accuracy and F1 scores when the original ratio is not available. Thus, it suggests that a higher masking ratio facilitates unsupervised representation learning in partially observable environments. If we were to choose between decreasing or increasing the ratio with the same amount in pretraining, increasing the number could be a reasonable choice.

\section{Conclusion}

We propose MST-DIM in this paper to deal with partially observable environments through pretraining. MST-DIM is a contrastive method based on an estimate of mutual information bound and uses masking in unsupervised pretraining to ensure the agent can learn reliable latent generative factors. Experiments are conducted using a benchmark of unsupervised learning on the annotated interface of Atari 2600 games. MST-DIM shows the benefit of using unsupervised representation learning in partially observable environments by achieving higher accuracy and F1 scores than ST-DIM and supervised learning.

For future work, it would be interesting to directly apply MST-DIM to RL environments and evaluate its performance using RL baselines. Designing an auxiliary contrastive loss in RL is typical, but the implementation details can vary among different research. Exploiting the weight initialization of pretrained representation encoders that resembles more to CV probing tasks can also be an intriguing topic.

Recognizing small objects can be a challenging task in partially observable environments. ST-DIM deploys the local-local contrastive task to reliably learn the representations of small objects. However, it still suffers from information loss and obtains worse results in partially observable environments. MST-DIM deals with this issue and achieves scores close to ST-DIM under the fully observable setup. It still remains to be studied how to recognize small objects where the environment is almost invisible, and information loss is severe. 

\subsubsection{Acknowledgements} This work was performed on the [ML node] resource, owned by the University of Oslo, and operated by the Department for Research Computing at USIT, the University of Oslo IT-department. http://www.hpc.uio.no/

{\scriptsize
\bibliographystyle{splncs04}
\bibliography{main}

\begin{thebibliography}{10}
\providecommand{\url}[1]{\texttt{#1}}
\providecommand{\urlprefix}{URL }
\providecommand{\doi}[1]{https://doi.org/#1}

\bibitem{anand2019unsupervised}
Anand, A., Racah, E., Ozair, S., Bengio, Y., C{\^o}t{\'e}, M.A., Hjelm, R.D.:
  Unsupervised state representation learning in atari. Advances in neural
  information processing systems  \textbf{32} (2019)

\bibitem{burda2018large}
Burda, Y., Edwards, H., Pathak, D., Storkey, A., Darrell, T., Efros, A.A.:
  Large-scale study of curiosity-driven learning. arXiv preprint
  arXiv:1808.04355  (2018)

\bibitem{cassandra1994acting}
Cassandra, A.R., Kaelbling, L.P., Littman, M.L.: Acting optimally in partially
  observable stochastic domains. In: Aaai. vol.~94, pp. 1023--1028 (1994)

\bibitem{chen2020simple}
Chen, T., Kornblith, S., Norouzi, M., Hinton, G.: A simple framework for
  contrastive learning of visual representations. In: International conference
  on machine learning. pp. 1597--1607. PMLR (2020)

\bibitem{chen2021empirical}
Chen, X., Xie, S., He, K.: An empirical study of training self-supervised
  vision transformers. In: Proceedings of the IEEE/CVF International Conference
  on Computer Vision. pp. 9640--9649 (2021)

\bibitem{devlin2018bert}
Devlin, J., Chang, M.W., Lee, K., Toutanova, K.: Bert: Pre-training of deep
  bidirectional transformers for language understanding. arXiv preprint
  arXiv:1810.04805  (2018)

\bibitem{doersch2017multi}
Doersch, C., Zisserman, A.: Multi-task self-supervised visual learning. In:
  Proceedings of the IEEE international conference on computer vision. pp.
  2051--2060 (2017)

\bibitem{dosovitskiy2020image}
Dosovitskiy, A., Beyer, L., Kolesnikov, A., Weissenborn, D., Zhai, X.,
  Unterthiner, T., Dehghani, M., Minderer, M., Heigold, G., Gelly, S., et~al.:
  An image is worth 16x16 words: Transformers for image recognition at scale.
  arXiv preprint arXiv:2010.11929  (2020)

\bibitem{gregor2015draw}
Gregor, K., Danihelka, I., Graves, A., Rezende, D., Wierstra, D.: Draw: A
  recurrent neural network for image generation. In: International conference
  on machine learning. pp. 1462--1471. PMLR (2015)

\bibitem{guo2020bootstrap}
Guo, Z.D., Pires, B.A., Piot, B., Grill, J.B., Altch{\'e}, F., Munos, R., Azar,
  M.G.: Bootstrap latent-predictive representations for multitask reinforcement
  learning. In: International Conference on Machine Learning. pp. 3875--3886.
  PMLR (2020)

\bibitem{he2020momentum}
He, K., Fan, H., Wu, Y., Xie, S., Girshick, R.: Momentum contrast for
  unsupervised visual representation learning. In: Proceedings of the IEEE/CVF
  conference on computer vision and pattern recognition. pp. 9729--9738 (2020)

\bibitem{hjelm2018learning}
Hjelm, R.D., Fedorov, A., Lavoie-Marchildon, S., Grewal, K., Bachman, P.,
  Trischler, A., Bengio, Y.: Learning deep representations by mutual
  information estimation and maximization. arXiv preprint arXiv:1808.06670
  (2018)

\bibitem{jonschkowski2015learning}
Jonschkowski, R., Brock, O.: Learning state representations with robotic
  priors. Autonomous Robots  \textbf{39}(3),  407--428 (2015)

\bibitem{kingma2013auto}
Kingma, D.P., Welling, M.: Auto-encoding variational bayes. arXiv preprint
  arXiv:1312.6114  (2013)

\bibitem{kolesnikov2019revisiting}
Kolesnikov, A., Zhai, X., Beyer, L.: Revisiting self-supervised visual
  representation learning. In: Proceedings of the IEEE/CVF conference on
  computer vision and pattern recognition. pp. 1920--1929 (2019)

\bibitem{laskin2020curl}
Laskin, M., Srinivas, A., Abbeel, P.: Curl: Contrastive unsupervised
  representations for reinforcement learning. In: International Conference on
  Machine Learning. pp. 5639--5650. PMLR (2020)

\bibitem{lee2020predictive}
Lee, K.H., Fischer, I., Liu, A., Guo, Y., Lee, H., Canny, J., Guadarrama, S.:
  Predictive information accelerates learning in rl. Advances in Neural
  Information Processing Systems  \textbf{33},  11890--11901 (2020)

\bibitem{lesort2018state}
Lesort, T., D{\'\i}az-Rodr{\'\i}guez, N., Goudou, J.F., Filliat, D.: State
  representation learning for control: An overview. Neural Networks
  \textbf{108},  379--392 (2018)

\bibitem{mnih2013playing}
Mnih, V., Kavukcuoglu, K., Silver, D., Graves, A., Antonoglou, I., Wierstra,
  D., Riedmiller, M.: Playing atari with deep reinforcement learning (2013)

\bibitem{oord2018representation}
Oord, A.v.d., Li, Y., Vinyals, O.: Representation learning with contrastive
  predictive coding. arXiv preprint arXiv:1807.03748  (2018)

\bibitem{puterman1990markov}
Puterman, M.L.: Markov decision processes. Handbooks in operations research and
  management science  \textbf{2},  331--434 (1990)

\bibitem{stooke2021decoupling}
Stooke, A., Lee, K., Abbeel, P., Laskin, M.: Decoupling representation learning
  from reinforcement learning. In: International Conference on Machine
  Learning. pp. 9870--9879. PMLR (2021)

\bibitem{sutton2018reinforcement}
Sutton, R.S., Barto, A.G.: Reinforcement learning: An introduction. MIT press
  (2018)

\bibitem{taylor2008playing}
Taylor, L.N., Whalen, Z.: Playing the past: History and nostalgia in video
  games. JSTOR (2008)

\bibitem{ulyanov2018deep}
Ulyanov, D., Vedaldi, A., Lempitsky, V.: Deep image prior. In: Proceedings of
  the IEEE conference on computer vision and pattern recognition. pp.
  9446--9454 (2018)

\bibitem{watkins1992q}
Watkins, C.J., Dayan, P.: Q-learning. Machine learning  \textbf{8}(3-4),
  279--292 (1992)

\bibitem{zhu2022masked}
Zhu, J., Xia, Y., Wu, L., Deng, J., Zhou, W., Qin, T., Liu, T.Y., Li, H.:
  Masked contrastive representation learning for reinforcement learning. IEEE
  Transactions on Pattern Analysis and Machine Intelligence  (2022)

\end{thebibliography}

\end{document}